# On the equivalence between Kolmogorov-Smirnov and ROC curve metrics for binary classification

Paulo J. L. Adeodato                     Sílvio B. Melo
Universidade Federal de Pernambuco, Centro de Informática
Av. Jornalista Aníbal Fernandes, Cidade Universitária
Recife-PE, Brazil,
+55 81 2126 8430

pjla@cin.ufpe.br                     sbm@cin.ufpe.br

## ABSTRACT
Binary decisions are very common in artificial intelligence. Applying a threshold on the continuous score gives the human decider the power to control the operating point to separate the two classes. The classifier's discriminating power is measured along the continuous range of the score by the Area Under the ROC curve (AUC_ROC) in most application fields. Only finances uses the poor single point metric maximum Kolmogorov-Smirnov (KS) distance. This paper proposes the Area Under the KS curve (AUC_KS) for performance assessment and proves AUC_ROC = 0.5 + AUC_KS, as a simpler way to calculate the AUC_ROC. That is even more important for ROC averaging in ensembles of classifiers or $n$-fold cross-validation. The proof is geometrically inspired on rotating all KS curve to make it lie on the top of the ROC chance diagonal. On the practical side, the independent variable on the abscissa on the KS curve simplifies the calculation of the AUC_ROC. On the theoretical side, this research gives insights on probabilistic interpretations of classifiers assessment and integrates the existing body of knowledge of the information theoretical ROC approach with the proposed statistical approach based on the thoroughly known KS distribution.

## ACM-Classification
I.5 PATTERN RECOGNITION, I.5.2 Design Methodology, Classifier design and evaluation;
I.2 ARTIFICIAL INTELLIGENCE

## 1. INTRODUCTION
In business, one of the most common decision strategies for selecting the eligible individuals or objects for an action is ranking them according to a classification score and choosing those above a pre-defined threshold [1]. That is used in applications such as staff selection, fraud detection [2], and resources allocation in public policies, for instance.

This score is computed by either weighing a set of variables based on human defined parameters or by applying a function learned by a classification algorithm from a set of data with binary labels as desired responses, according to specific optimization criteria. The mapping of a multidimensional space into a scalar variable (score) is fundamental for giving the human decisor the power to control the decision by simply setting a threshold in a continuous fashion. And that is applicable to both human weighed and data learned responses provided that these systems produce a score. This thresholding strategy can be used for decisions based not only on the classifier's technical performance but also on the business Key Performance Indicators (KPIs) [1].

As for general-purpose applications, there is not yet a specific use defined for a classifier in a decision support system, its operating point (decision threshold) cannot be pre-specified. In such scenario, the performance assessment metrics should focus on a feature of the classifier itself, without any assumption on the operating point. Therefore, the classifier performance evaluation should consider all the score range instead of a single point, unless specified by the application domain requirements.

That is probably the reason why the area under the curve of the Receiver Operating Characteristics (AUC_ROC) [1] became so popular among scientists for binary classification quality assessment, instead of confusion matrix, error rates, maximum Vertical Distance of the ROC curve, maximum Kolmogorov-Smirnov vertical difference and all other single point specific metrics.

Despite being used in financial applications [3] as dissimilarity metric [4] to assess continuous score classifier performance, the maximum Kolmogorov-Smirnov vertical difference (Max_KS2) is constrained to a specific point which rarely fits the business requirements. Nevertheless it is an important metric with consolidated statistical knowledge for hypothesis testing [5].

This paper puts forward the Area Under the Curve of the Kolmogorov-Smirnov (AUC_KS) distribution as a novel metric for binary classification performance assessment based on continuous score ranking. The paper also associates it with the ROC curve [6] by proving that AUC_KS = AUC_ROC − 0.5, with some advantages over one the most widely accepted performance metrics for binary classifiers (AUC_ROC) [1,7].

This paper is organized in three more sections. Section 2 presents the ROC and the KS concepts and curves for binary classification assessment. Section 3 presents the proof of the equivalence of the areas. Section 4 concludes the paper discussing some impacts of such formal statistical background for the interpretation of binary classification and its applications and further theoretical advances.

## 2. BINARY CLASSIFICATION PERFORMANCE METRICS
For a general-purpose binary classifier to be used in a decision support system, its response should be continuous to give the human decider the power to control the impacts of the decision. Classifiers that produce "hard" decisions by already presenting the predicted class are not the focus of this research for their lack of control flexibility. The classifiers of interest here map a multidimensional input space into a scalar (the score) over which the decider sets the decision threshold to create the two classes based on the business KPIs. For this study, the quality of the classifier is assessed by comparing the predicted class against the true class for each pattern of a test sample for all potential decision thresholds.

Consequently, the performance assessment metrics should also cover at least a range of this scalar in the region of interest for decision making [1]. Threshold specific metrics such as error rates are not adequate to assess the quality of these flexible classifiers. This paper is focused on area-based metrics such as AUC_ROC, Gini coefficient and the proposed AUC_KS, which measure their performance by integrating the impact of the classifier over the score range. These are all related and this work is particularly focused on the equivalence between the areas under the ROC curve and the Kolmogorov-Smirnov statistical distribution curve.

### 2.1 ROC Curve

The ROC Curve [6] is a non-parametric performance assessment tool that represents the compromise between the true positive rate (TP) and the false positive rate (FP) of example classifications based on a continuous output along all its possible decision threshold values (the score). In medical environment, the ROC curve equivalently expresses the compromise between sensitivity and specificity (actually, 1- specificity). Despite of the ROC curve being able to provide single point and area-based performance metrics, this paper will focus on the latter (AUC_ROC), which is independent of the operating point or any other specific point. It is a performance indicator valid for assessing the performance throughout the whole continuous range of scores [6-8]. Considering binary decision on continuous range, the bigger the AUC_ROC, the closer the system is to the ideal classifier (AUC_ROC=1). If the ROC curve of a classifier appears above that of another classifier along the entire domain of variation, the former system is better than the latter.

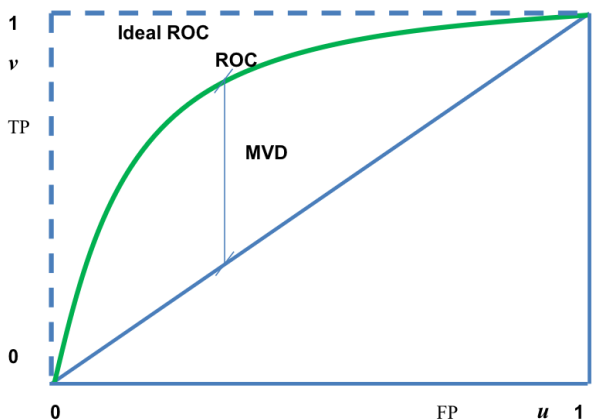

**Figure 1. The ROC curve.**

Figure 1 shows the ROC curve with its most important features. The ideal classifier is shown in dashed lines and it defines an isosceles triangle of area 0.5 with the chance diagonal that represents randomly sorted examples from the test sample. The figure also depicts the Maximum Vertical Distance (MVD) from the ROC curve to the chance diagonal that is used as single point metric of performance assessment, which has already been proven equivalent to the Maximum Kolmogorov-Smirnov difference (Max_KS) [7].

### 2.2 Kolmogorov-Smirnov Distribution

The Kolmogorov-Smirnov distribution has been originally conceived as an adherence hypothesis test for distribution fitting to data [5]. In binary classification problems, it has been used as dissimilarity metric for assessing the classifier´s discriminant power measuring the distance that its score produces between the cumulative distribution functions (CDFs) of the two data classes [4,7], known as KS2 for this purpose (two samples). The usual metric for both purposes is the maximum vertical difference between the CDFs (Max_KS), which is invariant to score range and scale making it suitable for classifiers comparisons. However, that is an assessment at a single operating point that, in general, does not suit the applications´ needs.

So the first author has proposed [9] an adaptation of the Kolmogorov-Smirnov metric, which is invariant to score scale and range, allowing its application for binary classifiers' performance assessment over the whole score range, independent of operating point. The examples are sorted by their score values and divided in quantiles thus making the Area Under the Kolmogorov-Smirnov curve (AUC_KS) a robust metric for performance assessment. Both KS-based metrics have also been applied to assess the quality of continuous input variables [10].

The same author has also proposed an algorithm for input variable transformation optimization [11] based on the Max_KS2 metric. The method consists of dividing the variable range in segments of monotonic variation of the KS curve and reordering these segments to group all increasing ones on the left hand side to reach the maximum KS followed by the group of all decreasing ones on the right hand side until reaching zero (0), thus maximizing the overall Max_KS2 for the transformed variable.

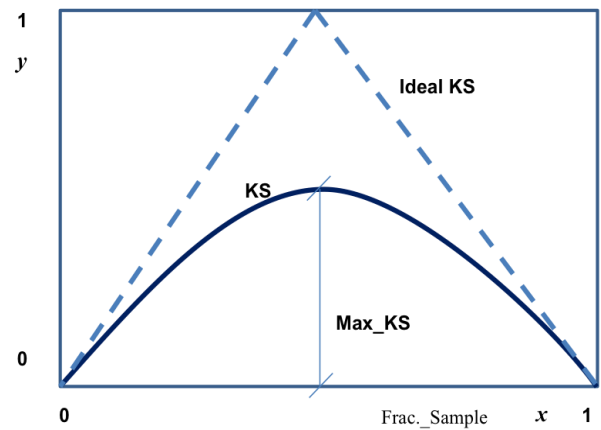

**Figure 2. The Kolmogorov-Smirnov curve.**

This paper formally proves that the AUC_KS metric for the classifier´s performance assessment is equivalent to the

AUC_ROC. In fact, this paper proves that AUC_ROC= 0.5+AUC_KS.

## 3. THE PROOF OF EQUIVALENCE OF THE AREAS UNDER THE CURVES

### 3.1 Inspiration

The research on the equivalence of areas was inspired by both the effectiveness of the metrics in binary decision problems and their geometrical similarity considering the rotation and axis expansion and reduction as illustrated in Figure 3 below.

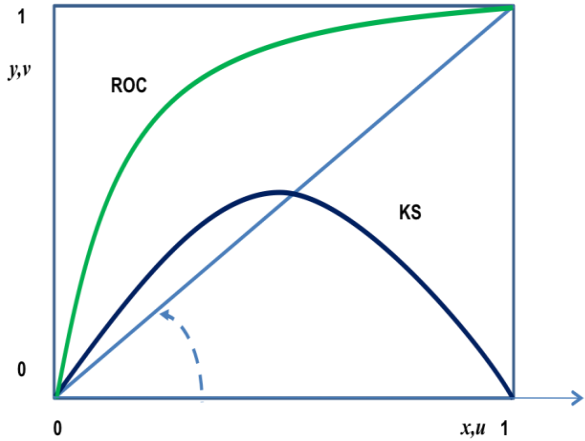

**Figure 3. Graphical transformation between ROC and KS**

The conceptual equivalence between the abscissa in the KS curve (*x* axis) and the chance diagonal in the ROC curve was the basis of reasoning. Also, an important aspect that helped the analysis was the limit case of the ideal classifier. That is the classifier that sets all examples from the target class apart from those from the complementary class. It is clear that the AUC_ROC is equal to one, while the AUC_KS is defined by a triangle that reaches the maximum height (1) making the AUC_KS=0.5. However, Figure 2 shows that the AUC_KS is defined by the non-isosceles triangle whose asymmetry depends on the unbalance between the classes. The multiplicity of KS curves for ideal classifiers mapped into a single ROC curve for all classifiers seemed an obstacle to the equivalence. All *n*-2 possible optimal classifiers ranging from having 1 to *n*-1 examples of the target class in a sample of size *n* would be mapped on the same ROC curve.

### 3.2 Proof of Equivalence

Let us first define the quantities and relative frequencies needed for building up the ROC and the KS curves at each potential decision point *i*:

- $n$ is the sample size
- $n_T$ is the number of examples of the target class
- $n_{\bar{T}}$ is the number of examples of the complementary class
- $n_{T_i}$ is the number of examples of the target class up to the *i*-th example
- $n_{\bar{T}_i}$ is the number of examples of the complementary class up to the *i*-th example
- $x, y$ are the axes of the KS curve
- $u, v$ are the axes of the ROC curve

In the construction method of calculation sheets for plotting both the Kolmogorov-Smirnov and the Receiver Operating Characteristics, after sorting the target labels according to the propensity score, these performance indicators are a sequence of simple operations performed on the relative frequencies of the target class, the complementary class and the population.

By definition of a ROC curve: $u = \frac{n_{\bar{T}_i}}{n_{\bar{T}}}$ and $v = \frac{n_{T_i}}{n_T}$, and by definition of a KS curve: $x = \frac{n_{T_i}}{n} + \frac{n_{\bar{T}_i}}{n}$ (abscissa) and $y = \frac{n_{T_i}}{n_T} - \frac{n_{\bar{T}_i}}{n_{\bar{T}}}$ (ordinate).

The most parsimonious assumption about this curve equivalence is one being a linear mapping of the other. To determine this 2D transformation it is required only to provide the image of two independent vectors. Clearly, one of these correspondences comes from the identification of the curves' endpoints; therefore, the vector (*x*=1,*y*=0) should be mapped into the vector (*u*=1,*v*=1). The other correspondence can be deduced from identifying the curves from ideal classifiers in both systems, since they establish the boundaries inside which a curve representing any other classifier must lie. In the case of the ROC representation, the ideal classifier curve is formed by the line segments from (0, 0) to (0,1), and from (0,1) to (1,1), following the order of descending scores, showed as dashed lines in Figure 1. The corresponding curve in the KS representation is formed by line segments from (0, 0) to ($\frac{n_T}{n}$, 1), and from ($\frac{n_T}{n}$, 1) to (1, 0), showed as dashed lines in Figure 2. Therefore, the vector ($x = \frac{n_T}{n}$, $y = 1$) should be mapped into the vector (*u*=0,*v*=1). From these correspondences, by using basic linear algebra, it can be shown that the resulting transformation is given as:

$$T\begin{bmatrix}x\\y\end{bmatrix} = \begin{bmatrix} 1 & -\frac{n_T}{n} \\ 1 & \left(1 - \frac{n_T}{n}\right) \end{bmatrix} \cdot \begin{bmatrix}x\\y\end{bmatrix}$$

This transformation can be seen as the composition of a rotation of $\pi/4$ rad (counterclockwise) followed by an anisotropic scaling of $\sqrt{2}$ in the *x* direction and of $1/\sqrt{2}$ in the *y* direction, followed by a shearing in the *x* direction with a factor of $\frac{1}{2} - \frac{n_T}{n}$.

It should be noticed that area is invariant under this transformation, since its determinant equals one [12]:

$$det \begin{vmatrix} 1 & -\frac{n_T}{n} \\ 1 & \left(1 - \frac{n_T}{n}\right) \end{vmatrix} = \left(1 - \frac{n_T}{n}\right) - \left(-\frac{n_T}{n}\right) = 1$$

This matrix, however, is not orthogonal, since its columns are not unitary. Therefore, lengths and angles are not invariant under this transformation (as it should be expected when anisotropic scaling and shearing are involved) [12].

By substituting the KS expressions of *x* and *y* in the matrix definition of *T* results in:

$$T\begin{bmatrix}\frac{n_{T_i}}{n}+\frac{n_{\bar{T}_i}}{n}\\ \frac{n_{T_i}}{n_T}-\frac{n_{\bar{T}_i}}{n_{\bar{T}}}\end{bmatrix}=\begin{bmatrix}\frac{n_{T_i}}{n}+\frac{n_{\bar{T}_i}}{n}-\frac{n_T}{n}\left(\frac{n_{T_i}}{n_T}-\frac{n_{\bar{T}_i}}{n_{\bar{T}}}\right)\\ \frac{n_{T_i}}{n}+\frac{n_{\bar{T}_i}}{n}+\left(1-\frac{n_T}{n}\right)\left(\frac{n_{T_i}}{n_T}-\frac{n_{\bar{T}_i}}{n_{\bar{T}}}\right)\end{bmatrix}$$

$$=\begin{bmatrix}\frac{n_{\bar{T}_i}}{n}+\frac{n_T\cdot n_{\bar{T}_i}}{n\cdot n_{\bar{T}}}\\ \frac{n_{T_i}}{n}+\frac{n_{\bar{T}_i}}{n}+\frac{n_{T_i}}{n_T}-\frac{n_{T_i}}{n}-\frac{n_{\bar{T}_i}}{n_{\bar{T}}}+\frac{n_T\cdot n_{\bar{T}_i}}{n\cdot n_{\bar{T}}}\end{bmatrix}$$

$$=\begin{bmatrix}\frac{n_{\bar{T}}\cdot n_{\bar{T}_i}+n_T\cdot n_{\bar{T}_i}}{n\cdot n_{\bar{T}}}\\ \frac{n_{\bar{T}_i}}{n}+\frac{n_{T_i}}{n_T}-\frac{n_{\bar{T}_i}}{n_{\bar{T}}}+\frac{n_T\cdot n_{\bar{T}_i}}{n\cdot n_{\bar{T}}}\end{bmatrix}$$

$$=\begin{bmatrix}\frac{(n_T+n_{\bar{T}})\cdot n_{\bar{T}_i}}{n\cdot n_{\bar{T}}}\\ \frac{n_{T_i}}{n_T}+\frac{n_{\bar{T}}\cdot n_{\bar{T}_i}}{n\cdot n_{\bar{T}}}-\frac{n\cdot n_{\bar{T}_i}}{n\cdot n_{\bar{T}}}+\frac{n_T\cdot n_{\bar{T}_i}}{n\cdot n_{\bar{T}}}\end{bmatrix}$$

$$=\begin{bmatrix}\frac{n\cdot n_{\bar{T}_i}}{n\cdot n_{\bar{T}}}\\ \frac{n_{T_i}}{n_T}+\frac{(n_T+n_{\bar{T}})\cdot n_{\bar{T}_i}-n\cdot n_{\bar{T}_i}}{n\cdot n_{\bar{T}}}\end{bmatrix}$$

$$=\begin{bmatrix}\frac{n_{\bar{T}_i}}{n_{\bar{T}}}\\ \frac{n_{T_i}}{n_T}+\frac{n\cdot n_{\bar{T}_i}-n\cdot n_{\bar{T}_i}}{n\cdot n_{\bar{T}}}\end{bmatrix}$$

$$=\begin{bmatrix}\frac{n_{\bar{T}_i}}{n_{\bar{T}}}\\ \frac{n_{T_i}}{n_T}\end{bmatrix}=\begin{bmatrix}u\\ v\end{bmatrix}$$

which represents both the False Positive rate coordinate (abscissa) and the True Positive rate coordinate of the ROC curve. In fact, not only there is equivalence between the axes of the ROC and KS curves but also they define equivalent areas. That is a consequence of having their limiting curves (for ideal classifiers), and their baselines, namely the *x* axis in the case of KS and the diagonal in the case of ROC, correspondent through the employed linear transformation, together with the fact that the area is invariant under the linear mapping.

Therefore, the Area Under the Kolmogorov-Smirnov curve (AUC_KS) is equal to the area of the ROC curve above its diagonal from (0, 0) to (1, 1) which defines a triangle of area equal to 0.5. In other words, the AUC_ROC = 0.5 + AUC_KS. That can be visually observed by the change of coordinates depicted in Figure 3.

Experimental assessment carried out on large actual datasets have shown that the equality AUC_ROC=0.5+AUC_KS holds in a precision of all eight decimals used in the experiments, as expected.

### 3.3 Equivalence for the Single Point Metrics

Considering the same transformation matrix from the previous subsection, the single point metric equivalent to the Max_KS2 in the ROC curve would be the maximum distance from the chance diagonal. However, as the distances orthogonal to the *x*-axis are mapped with a shrinking factor $\sqrt{2}$, the Max_KS2 projection is reduced to Max_KS2/$\sqrt{2}$ on the ROC curve. Conceptually, this point metrics, in principle, assesses maximum dissimilarity to the random classifier, which randomly sorts examples from the test sample.

Krzanowski and Hand have already proved the equivalence in value of the single point metrics [7] MVD (or the Youden Index) on the ROC curve with the Max_KS2 difference on the KS curve. However, strictly speaking, that is not conceptually equivalent to the Max_KS2 as a single point metric.

To illustrate that, let us consider a set of nine examples, which have been sorted according to a classifier score resulting in the binary sequence {1,0,0,1,0,1,0,0,0}.

The Max_KS2 metric is equal to 0.5 and occurs when the threshold is set after the 6[th] example as shown in Figure 4 below.

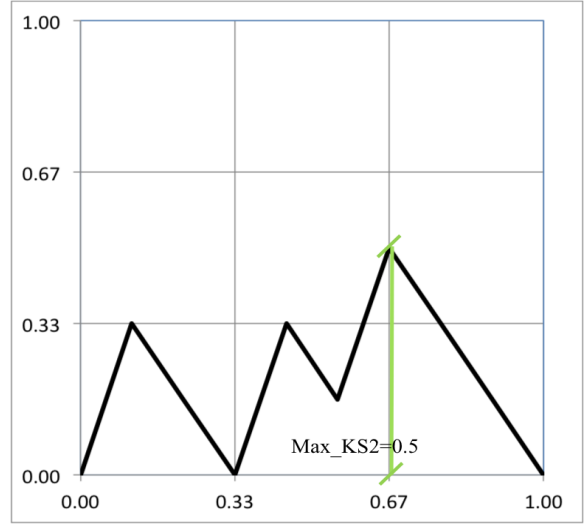

**Figure 4. The Kolmogorov-Smirnov curve of the example.**

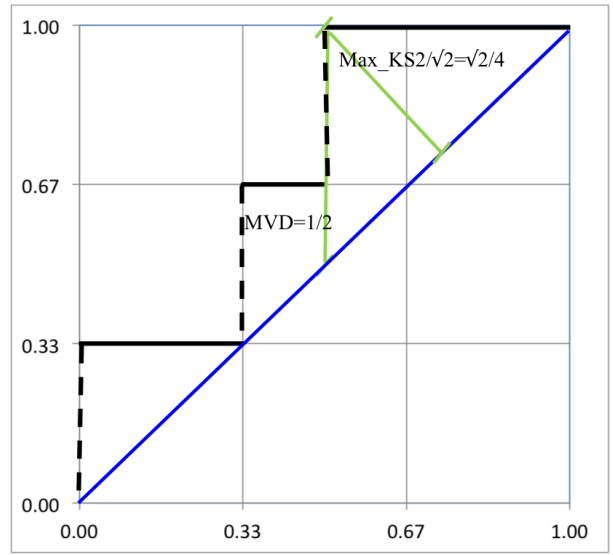

**Figure 5. The ROC curve of the example.**

Figure 5 shows the ROC curve of the example above and both single point metrics, namely, the MVD and the Max_KS2 projection. It is important to emphasize that both metrics values occur at the same threshold set after the 6[th] example. As MVD is equivalent in value to Max_KS2, it is larger than Max_KS2 projection by a factor of $\sqrt{2}$.

## 4. CONCLUSIONS

This paper has demonstrated the equivalence between the Area Under the Curve of the Receiver Operating Characteristics (AUC_ROC) curve and the Area Under the Curve of the Kolmogorov-Smirnov distribution (AUC_KS). In fact, the paper has proved that AUC_ROC = 0.5 + AUC_KS.

This result is very important once that it has put at the scientists' disposal for performance assessment of binary classification all the consolidated statistical knowledge on the Kolmogorov-Smirnov distribution available since 1933 [4]. That is over 20 years before the ROC analysis had been used in telecommunications [14] and much longer before its debut in artificial intelligence. This paper opens up a perspective of integrating the statistical and the information theoretical approaches.

Despite not having a formal proof before, the first author had already detected the numerical equivalence between these metrics and had been applying the Kolmogorov-Smirnov distribution as optimization criterion for embedding statistical knowledge in variables transformations [10]. With the equivalence proved, one just needs to optimize the AUC_KS whenever the classifier´s performance is measured by the AUC_ROC.

In practical applications such as data mining, this theoretical result also simplifies the averaging of ensembles of classifiers to estimate their overall performance. For the ROC, one has to choose either vertical averaging or threshold averaging [6]. The former calculation involves interpolation and has a higher level of imprecision than the latter. Threshold averaging involves the return to the scores for thresholding at specific sample quantiles, and the calculation of error-bars in both coordinates at each point, as detailed in reference [6]. The return to the scores is inherent to the ROC curve being a parametric curve with the classifiers' scores not visible. The KS curve has the score rank explicit as independent variable on the abscissa and has its error-bars only on the vertical axis (ordinate). One just needs to calculate the KS and error-bars at specific quantiles and apply the proposed transformation to get the error-bars projected in both ROC coordinates. That is a much simpler, yet precise, approach than those analysed in [15].

Furthermore, in decision support systems (DSSs), the managers need to define the decision threshold on the score (operating point) to produce the binary decision. That score is directly available in the Kolmogorov-Smirnov distribution. The direct use of the score is important not only for assessing the technical separability of the classifier but also for simulating the operating point on the business Key Performance Indicators (KPIs). An interactive environment such as [16] would be very effective for decision making on the control parameter.

This research has completed the equivalence established by Krzanowski and Hand between ROC and KS curves for single point metrics, extending it to the areas under the curves, which are the most important metrics for binary classification. It has brought a new perspective to the field for interpreting binary decisions and its impacts go much beyond this paper. One possible impact of using the AUC_KS metrics could be the definition of confidence intervals for the performance assessment of the classifiers without the need to resample data. The Kolmogorov-Smirnov statistics however has been developed only for the maximum vertical difference between the CDFs [4] so far. That is a challenge left for statisticians. In other words, this is just the beginning; not the conclusion.